\def\year{2019}\relax
\documentclass[letterpaper]{article} %DO NOT CHANGE THIS
\usepackage{aaai19}  %Required
\usepackage{times}  %Required
\usepackage{helvet}  %Required
\usepackage{courier}  %Required
\usepackage{url}  %Required
\usepackage{graphicx}  %Required
\usepackage{amssymb}
\usepackage{mathtools}
\usepackage{multirow}
\begin{document}
\title{Explaining Deep Learning Models using Causal Inference}
\author{Paper ID: 5792}
\author{Tanmayee Narendra, Anush Sankaran, Deepak Vijaykeerthy, Senthil Mani
\\
IBM Research
\\
\{tanarend, anusank, deepakvij, sentmani\}@in.ibm.com
}

\maketitle
\begin{abstract}
Although deep learning models have been successfully applied to a variety of tasks, due to the millions of parameters, they are becoming increasingly opaque and complex. In order to establish trust for their widespread commercial use, it is important to formalize a principled framework to reason over these models. In this work, we use ideas from causal inference to describe a general framework to reason over CNN models. Specifically, we build a Structural Causal Model (SCM) as an abstraction over a specific aspect of the CNN. We also formulate a method to quantitatively rank the filters of a convolution layer according to their counterfactual importance. We illustrate our approach with popular CNN architectures such as LeNet5, VGG19, and ResNet32.  
\end{abstract}

\section{Introduction}

Machine learning has come a long way in the past decade with the dramatic development of Deep Neural Networks (DNNs), and cheap availability of high-end general-purpose GPUs (GPGPUs) for high-speed computation~\cite{chen2014fast}, ~\cite{raina2009large}. These advancements have considerably improved the state-of-the-art in speech recognition, computer vision, natural language processing, etc. This success comes at a cost - DNNs derive their expressive power by learning a complex higher dimensional representation of the input data. In turn, the task of understanding intricacies of the intermediate representations and feature spaces of DNNs has become notoriously difficult. In spite of their opacity, DNNs are being used to power several end-user facing applications such as image recognition, machine translation as well as sensitive applications such as autonomous navigation~\cite{alvinn}, malware detection~\cite{malware}, radiology~\cite{radiology}, etc. and they influence our day to day decisions. Due to these factors, it has become paramount that we develop a framework to understand the inner workings of these models. 

The need for frameworks for interpreting and explaining the inner working of these complex models hasn't gone unnoticed. Several approaches have been proposed in the recent past to understand the predictions output by DNNs. In particular, there has been significant progress in explaining the predictions of a Convolution Neural Network (CNN) in the image domain. These techniques work by generating saliency maps which indicate the relevance of pixels in the image to the output of CNN. \cite{selvaraju2017grad} \cite{zhou2016learning}

Even though the existing methods have significantly improved the interpretability of CNNs, they suffer from a critical shortcoming. These techniques don't help humans to reason how and why perturbations to the model's structure (for example removing a filter from the \textit{n}-th layer of the model) can impact the model's predictions. 

In this work, we borrow ideas from causal inference \cite{pearl2009causality} to provide a general abstraction of a DNN which allows for arbitrary causal interventions and queries and in turn offers an invaluable tool to quantitatively explain the importance and impact of the constituent components of a DNN on its performance. The ability to perform arbitrary causal interventions allows us to seamlessly capture the chain of causal effects from the input, to the filters (in turn the complex intermediate representations), to the DNN outputs. In addition to the above, the proposed abstraction will also allow us to answer a richer set of questions which the existing frameworks can't such as counterfactual queries. For instance, an end user would be able to ask ``What is the impact of the \textit{n}-th filter on the \textit{m}-th layer on the model's predictions?''.  Such an abstraction provides a powerful tool through which an end user can debug, test and analyze the properties of a deployed DNN.

In this paper, we focus on image-based models and study the significance of its components  (for instance, filters) based on a pre-decided metric, to illustrate the utility of the proposed framework. A standard way to measure the importance of model elements is through ablation testing. For a fixed data set, this involves removing specific component(s) from the CNN model, retraining the model on the same data set and testing the performance. The difference in the model performance is an indicator of the importance of the model components under study.  Although ablation testing is a simple and intuitive way of measuring the influence of model components, it is computationally expensive, and may not be feasible for complex models.

One of the key advantages of the proposed framework is that, as opposed to the above technique, our approach only requires a one-time construction of the causal abstraction of a CNN and does not require any re-training of the model to identify a filter's influence on the model's predictions. 

The following are the research contributions of this work - First, we describe a general framework to build a causal model as an abstraction to reason over a specific aspect of a DNN. Second, we provide a simple approach to validate the causal model learnt, to check if the causal model abstraction adequately captures the behaviour of the DNN model. Last, we describe a method to quantitatively rank DNN model components according to their importance. We illustrate our framework with well known image classification models such as VGG-19 \cite{vgg19}, ResNet-32 \cite{he2016deep} and LeNet-5 \cite{lecun1998gradient} with CIFAR10 \cite{krizhevsky09learningmultiple}.

\section{Background}
In this section, we will give a brief introduction to causality. A reader well versed in causal inference may skip this section.

\subsection{Causal Theory}
All explanations are arguably and inherently causal. Questions of the sort \textit{What if}, and \textit{Why} require an understanding of the underlying mechanisms of the system and the theory of causal inference is one method by which such questions might be answered.

Interpretability methods such as saliency maps \cite{zhou2016learning} \cite{selvaraju2017grad} only establish a correlation - while it is possible to say that  a particular image region is responsible for it to be correctly classified, it cannot say what would happen if a certain portion of the image was masked out (This is a \textit{What If} question.) 

In the statistical paradigm, the objective is to approximate the joint distribution from the data at hand, in order to answer relevant questions about new data. Statistical models allow us to answer questions related to prediction. A causal model, on the other hand, subsumes a statistical model, and contains additional structure that helps in answering questions about interventions and distribution changes. \cite{peters2017elements} 

Causal Models help in answering \textit{what if} kinds of questions. Consider the deep learning model as the system under study. Causal Models contain additional structure about the system that helps in answering questions of the kind - \textit{For this image, what if I black out this filter response to zero? Would my prediction still be correct?} and \textit{If I blacked out this filter completely, how much would my accuracy change?}. These kinds of what-if questions, which involve making changes to the system under study, can be answered through causal models.

Any causal model must be expressive enough to answer \textit{prediction}, \textit{intervention} and \textit{counterfactual} questions, which are in increasing order of difficulty. A \textit{prediction} question is essentially an association - it involves asking what would happen to a variable of interest, if the values of other variables were observed to be a given value. Prediction questions have been very well studied in Statistics and Machine Learning, and there are several successful methods for answering observational questions. \cite{pearl2009causality} \cite{peters2017elements} An \textit{intervention} question on the other hand, deals with answering what would happen to the variable of interest, if a subset of variables were \textit{set} to a particular value. A \textit{counterfactual }questions deals with answering \textit{What If} kinds of retrospective questions. \cite{pearl2009causality} 

Causality has a long history, and there are several formalisms such as Granger causality, Causal Bayesian Networks and Structural Causal Models. \cite{lattimore2018primer}. In this work, we consider the Structural Causal Model paradigm, which allows us to specify causal mechanisms by specifying a set of equations.

\subsubsection{Structural Causal Models (SCM)}
Consider a set of random variables $\mathbb{X}= \{ X_1, X_2, ..., X_n\}$. The Structural Causal Model for $\mathbb{X}$ is defined as the set of assignments
\[ X_i \coloneqq f_i (PA_i, N_i)\] where $f_i$ is a function, $PA_i \subset  \mathbb{X}$, ($X_i \notin \mathbb{X}$ ) and $N = \{N_i\}$ are jointly independent. $PA_i$ denotes the parents of $X_i$, and $N$ denotes the set of noise variables. Structural Causal Models are also called as Structural Equation Models. \cite{peters2017elements}

Every SCM is associated with a directed graph, where each edge can be colloquially interpreted as going from cause to effect. Using the equations of the SCM, the graph $G$ associated with the SCM is constructed as follows - construct a vertex for every $X_i \in \mathbb{X}$, and draw an edge from every vertex in $PA_i$ into $X_i$. In this work, we only consider Structural Causal Models whose graph $G$ is a Directed Acyclic Graph (DAG). \cite{peters2017elements} 

It is important to note that these equations are to be interpreted in as assignments, and are not bi-directional. 

\subsubsection{Estimating Interventions}
% change notation here (but how?) 
An intervention of setting a subset of variables $X \subset \mathbb{X}$ to a particular value is usually denoted as $do(X=x)$. 

The causal effect of $X$ on $Y$ can be estimated with 
$P(Y=y | do(X=x)) = \sum_{z} P(Y=y| X=x, PA=z) P(PA=z)$   
where $PA$ denotes the parents of $X$ in the associated graph, and $z$ ranges over the set of values taken by $PA$.   \cite{pearl2009causality} 

In this work however, we do not use this formulation because the high connectedness and the large number of variables in our causal graph makes estimating the marginal probabilities computationally expensive. Instead, for a fixed dataset $D$, we actually perform the intervention, and then estimate the expected value of $Y$ over $D$. This is explained in detail in Section 5.

\section{Existing Work}
There have been several interesting applications where building a causal model of a system can provide a principled approach to reasoning about it. In \cite{bottou2013counterfactual}, a causal model is built of a computational advertisement system. Here, the main quantity of interest is the advertisement clickthrough rate. The causal model was used to provide a principled approach to model counfounding, and a framework to inexpensively reason about the effect of interventions such as changing the ad-placement algorithm without conducting randomized trials.

Causal models have been used to isolate noise from the signals from the Kepler space telescope, and this has helped in the discovery of several previosuly undiscovered exoplanets. \cite{scholkopf2016modeling}

Modelling bias and fairness as a counterfactual is a new way to quanitify the bias of models. \cite{kusner2017counterfactual}

Ideas from causal inference have also been used in providing human understandable explanations in deep learning models. In \cite{harradon2018causal}, a method is developed to build a causal model over human understandable image based abstractions of the model, which helps in asking counterfactual questions. In \cite{lu2018gender}, a causal model is used to remove gender bias in NLP models. 

% write a summarize paragraph
% what is new in our work

Thus, causal inference has been applied to several domains with considerable success. To the best of our knowledge, this is the first work that uses causal inference to reason over DNN model components.
\section{Proposed Approach}
In this section, we explain how to build a causal model to reason over a deep learning model. To build a SCM of the DNN, first we build the DAG structure from the DNN. Second, we apply a suitable transformation, which captures the aspect of the DNN that we want to reason over. Last, we estimate the structural equations of the SCM. (See Figure \ref{fig:approach}) These steps are described in detail below. 

\subsection{Building the Structural Causal Model}
Let the deep learning model under study be $M_D$, and its corresponding structural causal model be $M_C$. We need to estimate $M_C$ from $M_D$.

\subsubsection{DAG Structure}
DNNs  have an inherent DAG structure, which we exploit to construct the skeleton DAG for the Structural Causal Model (SCM) $M_C$. This DAG structure can be visualized by considering every node in a neural network as a vertex in a graph, and every forward connection as a directed edge. This DAG in the SCM can be constructed at the granularity level to which the DNN model is to be studied. For example, suppose we wish to study a CNN at the granularity level of its filters. The SCM DAG is constructed such that for a given filter in the $i^{th}$ convolution layer, there is an incoming directed edge from from every filter in the $(i-1)^{th}$ layer. On the other hand, if we wanted to study a DNN at a layer level, the SCM DAG is constructed such that there is a directed edge from the $i^{th}$ layer to the $(i+1)^{th}$ layer. 

% add another granularity example

In this work, we will restrict ourselves to studying the CNN model at the level of its filters.

\subsubsection{Applying a Suitable Transformation}
The causal abstraction of the DNN is built to encapsulate some aspect of the underlying DNN model, over which reasoning, and \textit{what if} questions might be answered. To this effect, an appropriate transformation function $\phi$ must be selected, keeping in mind the nature of intervention questions that need to be answered from the causal model. For example, if it was required to answer questions about the variance of filter responses in convolution layers, a transformation that encodes variance must be formulated. The choice of $\phi$ is not restricted, and transformation functions can be arbitrarily complex.

However, there is one practical consideration - In a CNN, every filter response from a convolution layer is a matrix. We will need to convert this matrix to a real number in order to estimate the equations in the SCM. Hence, in this work, we consider only those transformation functions whose range is a real number. 

\begin{figure*}
\includegraphics[scale=0.6]{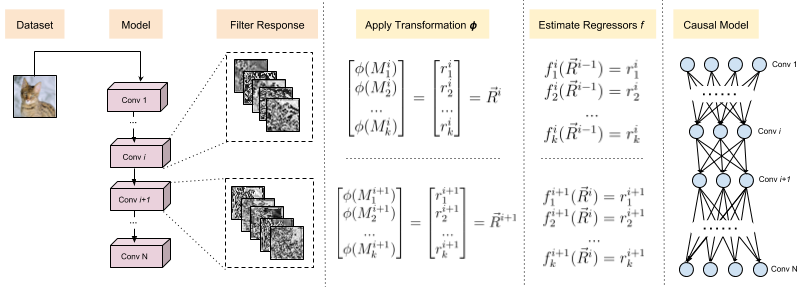}
\caption{Approach to build a SCM from a CNN \label{fig:approach}}
\end{figure*}

Let $\phi: \mathbb{R}^{p \times q} \rightarrow \mathbb{R} $ be a suitable transformation function, that captures that aspect of the filter that we wish to model in the SCM. If $M_j^i \in $ is the $j^{th}$ filter response of the $i^{th}$ layer, $\phi (M_j^i) = r_j^i$ is its transformation to a real number. (See Figure \ref{fig:approach})

\subsubsection{Learning the Equations of the SCM}
Now, we need to estimate the structural assignments, or structural equations of the causal model. 

For a Structural Causal Model whose causal graph is a DAG, the joint distribution can also be estimated with Causal Bayesian Networks. \cite{pearl2009causality} However, if we were to estimate the joint distribution as a set of conditional probability tables, these tables become prohibitively large owing to the high connectedness of the DAG constructed from the CNN model. To illustrate this point, consider the VGG 19 model. The last pooling layer of the third block has 256 filters, which are connected to 512 filters in the first convolutional layer of the fourth block. The size of the conditional probability table for one filter in the fourth block is $2^{256} \approx 10^{77}$ , which is prohibitively large. Hence, it is not possible to estimate the joint distribution with conditional probability tables.

Instead, we learn a function $f$ such that 
\[r = f(PA_r) \]
where $r$ is node in the causal DAG, and $PA_r$ denotes the parents of $r$. The function $f$ is arbitrary. This reduces to the standard regression problem. 

Let us see how this works with the SCM constructed from a CNN. Let the dataset be $D$. Consider the $i^{th}$ convolution layer of of the CNN, and let us assume that it consists of $k$ filters. For a given image $D_i \in D$, let the set of filter responses from $i^{th}$ convolution layer be $M = \{M^i_1, M^i_2, ..., M^i_k\}$.  

Let $\phi$ be the transformation function, then we have
\[
\begin{bmatrix}
\phi(M^i_1)\\ 
\phi(M^i_2)\\ 
...\\ 
\phi(M^i_k)
\end{bmatrix} = 
\begin{bmatrix}
r^i_1\\ 
r^i_2\\ 
...\\ 
r^i_k
\end{bmatrix} = 
\vec{R}^i
\]
Now, for the given layer $i$, we need to approximate the set of functions $F_i = \{f^i_1, f^i_2, ..., f^i_k\}$. Recollect that the causal DAG was constructed in such a way that for every filter in layer $i$ of the CNN, its parents are the set of all filters in the layer $i-1$. Hence, for the $j^{th}$ filter in the $i^{th}$ layer of the CNN, function in $f^i_j$ can be approximated as follows - 
\[r_j^i = f^i_j(\vec{R}^{i-1})\]
The set of all functions $f^i_j$, together with the causal DAG gives us the Structural Causal Model. 

% An overview of the entire approach is captured in Figure \ref{fig:approach}.
\section{Experimental Set Up}
In this section, we describe the details of the experiments conducted and summarize the results. In this work, we consider only Convolutional Neural Networks (CNN).

\subsection{Implementation Details}
We apply the framework described previously to CNN models, VGG19~\cite{vgg19}, ResNet32~\cite{he2016deep} and LeNet5~\cite{lecun1998gradient}, over CIFAR10 data set\cite{netzer2011svhn}. All models were trained using Keras with Tensorflow backend.  

\begin{table}
	\centering
	\begin{tabular}{ccc}
\hline
\textbf{Dataset}                   & \textbf{Model}     & \textbf{Accuracy} \\ \hline
\multirow{3}{*}{CIFAR10}  & LeNet 5   & 0.706     \\
                          & VGG19     & 0.91     \\
                          & ResNet 32 & 0.924     \\ \hline
% \multirow{3}{*}{CIFAR100} & LeNet 5   & 0        \\
%                           & VGG 19    & 0        \\
%                           & ResNet 32 & 0        \\ \hline
% \multirow{3}{*}{MNIST}    & LeNet 5   & 0        \\
%                           & VGG 19    & 0        \\
%                           & ResNet 32 & 0        \\ \hline
\end{tabular}
\caption{Performance of trained models \label{table:trained}}
\end{table}

Table \ref{table:trained} shows the performance of pre-trained models.

\begin{figure}[t]
\includegraphics[scale=0.55]{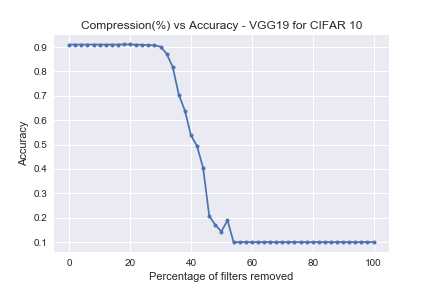}
\caption{Accuracy, when filters are removed in decreasing order of variance (VGG19 over CIFAR10) \label{fig:cifar10-vgg19}}
\end{figure}

\subsection{Building the Causal Model}
The causal DAG structure is inferred from the CNN model, and it remains fixed for a particular CNN model. 

There is some flexibility in the choice of the transformation function $\phi$ and the method of estimating the structural equations $F$. 

\subsubsection{Transformation $\phi$}
To build the SCM, we need to decide on an appropriate transformation $\phi$ , which captures some aspect of the filter that we want to reason about, and that can represent each filter of every CNN layer as a real number. 

Filters with low variance have been observed to contain more information than those with higher variance \cite{golub2018dropback}. The simplest representation is to convert every filter into a binary number, indicating whether or not each filter has high variance or not. An SCM constructed with this transformation would allow us to reason about the importance of filters, assuming that high/ low variance is a good measure of filter importance.

To this effect, we first consider the transformation 
\[\phi(M^i_j) = \left\{\begin{matrix}
1 & \left \| M^i_j \right \| < \mu^i_j + \sigma^i_j\\ 
0 & otherwise 
\end{matrix}\right.\]
where $\mu^i_j$ is the mean and $\sigma^i_j$ is the variance of the $j^{th}$  filter response from the $i^{th}$ layer $M^i_j$ over the dataset $D$. It is important to note that there is significant information loss with this transformation because, essentially we are converting a matrix to a bit. (If $M^i_j$ is of dimension $p \times q$ then $\log_{2}(pq) \rightarrow \log_{2} 2$)

\begin{figure}[t]
\includegraphics[scale=0.55]{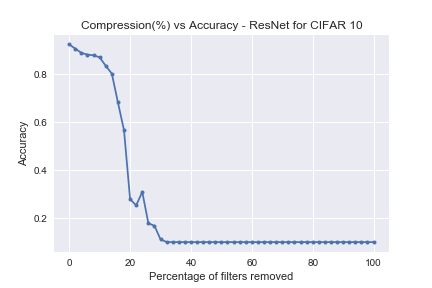}
\caption{Accuracy, when filters are removed in decreasing order of variance (ResNet 32 over CIFAR10) \label{fig:cifar10-resnet}}
\end{figure}

\begin{figure} [t]
\includegraphics[scale=0.55]{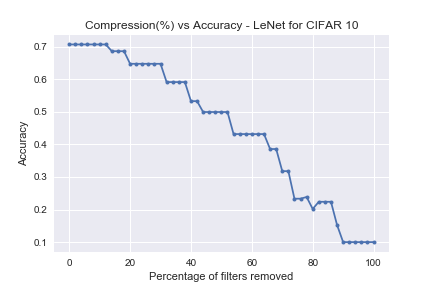}
\caption{Accuracy, when filters are removed in decreasing order of variance (LeNet 5 over CIFAR10) \label{fig:cifar10-lenet}}
\end{figure}

% can we replace table with figure
However, variance appears to be a good metric to reason about filter importance, as can be seen from Figure \ref{fig:cifar10-vgg19}, \ref{fig:cifar10-resnet}, \ref{fig:cifar10-lenet}.

We consider another transformation, which takes the Frobenius norm of $M^i_j$. 
\[\phi(M^i_j) = \left \| M^i_j \right \| =  \sqrt{ \sum_{p} \sum_{q} m_{pq}^2}\]
The information loss is lesser in this transformation. This transformation allows us to ask questions about the effect on the accuracy when intervening on a filter response and setting it to zero. 

Before fitting regression models on the transformed dataset, we augment the dataset by making random interventions within the model and observing the effect. In other words, for each set of filter responses from  a particular layer, we randomly zero out 10\% of the filters, and apply the transformation on the rest of the filter responses in the downstream layers of the model. This is to ensure that the SCM learns from intervention data as well. 

\begin{table}
\centering
\begin{tabular}{ccc}
\hline
\textbf{Model} & \textbf{Avg Accuracy} & \textbf{Min Accuracy} \\ \hline
Logistic Regression     & 0.926         & 0.526        \\
SVM & 0.848 & 0.659 \\
Random Forest      & \textbf{0.987}         & \textbf{0.722}       \\ \hline
\end{tabular}
\caption{Accuracy Metrics for SCM built over VGG19 (CIFAR10), using binary transformation \label{table:vgg19-regression-bin}}
\end{table}

\begin{table}
\centering
\begin{tabular}{ccc}
\hline
\textbf{Model} & \textbf{Avg MSE} & \textbf{Max MSE} \\ \hline
Linear     & \textbf{0.361}        &  \textbf{12.9 }       \\
Ridge      & 0.365            & 13.4       \\
LARS      & 73.1            & 2028.7      \\ \hline
\end{tabular}
\caption{Accuracy Metrics for SCM built over VGG19 32 (CIFAR10), Frobenius norm transformation \label{table:vgg19-regression-norm}}
\end{table}

\begin{table}
\centering
\begin{tabular}{ccc}
\hline
\textbf{Model} & \textbf{Avg MSE} & \textbf{Max MSE} \\ \hline
Linear     & 3.44  &  307.3       \\
Ridge      & \textbf{3.43}            & 298.3     \\
LASSO      & 3.56            &  \textbf{296.4}      \\\hline
\end{tabular}
\caption{Accuracy Metrics for SCM built over ResNet 32 (CIFAR10), Frobenius norm transformation \label{table:resnet-regression-norm}}
\end{table}

\begin{table}
\centering
\begin{tabular}{ccc}
\hline
\textbf{Model} & \textbf{Avg MSE} & \textbf{Max MSE} \\ \hline
Linear     & 22.2  &  \textbf{375.9}       \\
Ridge      & \textbf{22.1}            & 376.9     \\
LASSO      & 22.2           &  381.9    \\\hline
\end{tabular}
\caption{Accuracy Metrics for SCM built over LeNet5 (CIFAR10), Frobenius norm transformation \label{table:lenet-regression-norm}}
\end{table}

\subsubsection{Learning the Structural Equations}
Now, we estimate a set of regressions $F$, which gives us the SCM. For the binary transformation function, we used popular classification algorithms like Logistic Regression and Random Forest to learn the structural equations. See Table \ref{table:vgg19-regression-bin} for more details. The average accuracy metric is 98\% and is satisfactory.

For the Frobenius norm transformation, we used linear models to estimate the Structural Equations. Table \ref{table:vgg19-regression-norm}, Table \ref{table:resnet-regression-norm} and Table \ref{table:lenet-regression-norm} shows the average and maximum MSE for VGG-19, ResNet 32 and LeNet 5 architectures.

\section{Evaluation}
In this section, we describe a method by which we can check if the causal model built previously is representative enough of the CNN and adequately captures the information of each filter. This is done primarily to ensure that the transformation chosen adequately captures the behaviour of the CNN model. 

\subsection{Sanity Check of the Causal Model}
% rewrite this later

In order to check whether $\phi$ is an appropriate transformation, and whether the set of functions in the SCM $F = \{f_i^j\}$ adequately captures the model's performance, we run the dataset $D$  through $F$, and check if it gives a \textit{reasonable} accuracy. If the transformation function $\phi$ that was selected previously is not representative enough of the CNN, then the accuracy will be unsatisfactory. 

\subsection{Inference on the Causal Model}
Inference using the causal model can be done at two levels - we can answer questions relating to a particular sample from the dataset, and we can answer questions about the entire dataset. 

Let the sample be $D_i \in D$, let the subset of filters that we want to intervene on be denoted by $X$, and their values by $x$. The intervention of setting $X=x$ is denoted by $do(X=x)$. Let the filter whose behaviour we want to  observe after the intervention be denoted by $Y$. If we were interested in measuring the impact of an intervention on the model's performance, $Y$ would be chosen to be a filter from the last dense, classification layer of the CNN. It is important that $ Y \notin X$ . 

To estimate the effect of an intervention on $Y$, in the Structural Equations of the SCM, we  \textit{set} $X=x$ and estimate the value of $Y$. \cite{pearl2009causal}

To estimate the effect of an intervention on the dataset, we do the following - 
\[\mathbb{E}[Y|do(X=x)] = \frac {\sum_{D, M_C^{do(X=x)}} Y_i} {|D|}\]
In other words, for every sample $D_i \in D$, we estimate the value of $Y$ and take its average. 

\section{Results}
Based on the evaluation methodology, we present our results in this section.
\subsection{Sanity Check of the Causal Model}
Several useful questions that we want to answer about the deep learning model involve estimating the effect of an intervention on the model's performance. In order to check whether this is captured \textit{well enough} by the SCM, we measure the performance of the model over the dataset $D$ when just the SCM is used for prediction. 

In other words, every layer $i$ with $k$ filters in the CNN model is now associated with a set of Structural Equations $F^i=\{f^i_1, f^i_2, ..., f^i_k\}$ which we estimated by fitting regressions. Similar to a forward propagation through the CNN model, we pass the dataset $D$ through the set of Structural Equations and check if the accuracy is satisfactory.

\begin{table}
\begin{tabular}{ccc}
\hline
\textbf{Model}     & \textbf{SCM Accuracy} & \textbf{Model Accuracy} \\ \hline
VGG 19    & 0.902        &          0.91\\
LeNet 5   &  0.830            &        0.706  \\
ResNet 32 &  0.727           &        0.924  \\ \hline
\end{tabular}
\caption{SCM Accuracy for different models learnt for CIFAR10 \label{table:scm-accuracy}}
\end{table}

Table \ref{table:scm-accuracy} shows the accuracy of the SCM over CIFAR10 dataset for VGG19, LeNet 5 and ResNet 32 models. This accuracy is reasonable and satisfactory. 

We do not report accuracy metrics for the binary transformation, as the SCM model accuracy achieved is less than random choice. We hypothesize that although the average linear model accuracy is satisfactory, the information loss in the binary transformation is too large to give a satisfactory abstraction of the deep learning model.

\subsection{Inference on the Causal Model}
There are several kinds of intervention and counterfactual questions that might be answered from a well specified Structural Causal Model. However, in this preliminary work, we restrict ourselves to answering questions related to the importance of a feature for correct classification. 

We only consider the Frobenius norm transformation here, as the binary transformation fails to provide a satisfactory abstraction of the model. The intervention we seek to answer is as follows - what would be the effect on CNN model accuracy, if we set the Frobenius norm of a particular filter response to zero? This is the same as asking the effect on the model's performance, when one of the filters is set to zero. (It can be easily verified that the Frobenius norm of a matrix is zero, only when the individual elements of the matrix are also zero.)

\begin{table}[]
\centering
\begin{tabular}{ccc}
\hline
\textbf{Layer} & \textbf{Least Important} & \textbf{Most Important} \\ \hline
Conv2D 1       & 8, 55, 10                        & 15, 11, 20                      \\
Conv 2D 3      & 8, 39, 93                        & 56, 116, 1                      \\
Conv 2D 5      & 226, 76, 34                      & 211, 13, 88                     \\
Conv 2D 6      & 152, 164, 205                    & 71, 233, 175                    \\
Conv 2D 7      & 84, 73, 229                      & 177, 240, 16                    \\
Conv 2D 2      & 48, 4, 27                        & 23, 1, 49                       \\
Conv 2D 4      & 117, 115, 79                     & 50, 22, 39                      \\
Conv 2D 8      & 172, 102, 46                     & 81, 2, 64                       \\
Conv 2D 9      & 101, 317, 441                    & 309, 162, 373                   \\ \hline
\end{tabular}
\caption{Most Important and Least Important Filters for VGG 19 \label{table:vgg-importance}}
\end{table}

We use the accuracy drop of the model to provide an estimate of the most important filters in the VGG 19 model. Table \ref{table:vgg-importance} shows the least and most important filters for different convolution layers. The importance of filters of a particular layer are ranked in increasing order of the SCM accuracy, that is, a filter is deemed to be less important if an intervention of setting it to zero, gives a lesser drop in SCM accuracy.
\section{Discussion }

The framework described in this work is generic, and has the potential to be used for a wide variety of applications. 
\begin{enumerate}
    \item \textbf{Model Compression}: For compressing deep learning models, typically certain nodes or layers are removed and the model is retrained to verify lossy nature of the model compression. However, using the SCM the model's performance post compression could be predicted without the need for retraining.
    \item \textbf{Transfer Learning of Models:} It is a common practice to consider a model pre-trained using a large scale dataset such as ImageNet and then finetune the weights for the target task in hand. Using the learnt SCM model, we could potentially predict the performance of the original model on the target task, without the need for finetuning.
    \item \textbf{Model Accuracy Prediction:} For different datasets and tasks, typically an extensive hyperparameter search is performed. For every combination of hyperparemeter the deep learning model has to be retrained to study the effect on performance. However, using the learnt SCM model, the accuracy of a modified hyperparamter configuration could be predicted without the need for retraining. 
\end{enumerate}

% Transfer learning

% Trainless prediction of accuracy
\section{Conclusion and Future Work}
In this preliminary work, we provide a general framework for understanding a deep learning model by building an abstraction of a specific aspect of the DNN model using causal inference. In specific, we describe how a Structural Causal Model can be built, verified and used for inference. We illustrate the effectiveness of this method by using it to provide a ranking of filters of different layers in a CNN model. 

There are several avenues for improving the strength of the causal models learned. Instead of approximating the functions of the Structural Equations with linear models, it might be worthwhile to explore of using a model class that has more expressive power (capacity) than linear models, which might allow us to answer complex queries. A more representative causal model would potentially allow one to answer complex intervention and counterfactual questions.  Also, in this work, we have considered two basic transformation methods to build the causal model. There are potentially infinite possibilities of transformations to consider, which may prove to be more representative of the deep learning model. Lastly, the current formulation largely assumes that the dataset is fixed. This framework can be potentially extended to generalize over datasets.
\bibliography{ref}
\bibliographystyle{aaai}
\end{document}